\documentclass[journal]{IEEEtran}
\usepackage[utf8]{inputenc}
\usepackage{times,enumerate} 
\usepackage[usenames,dvipsnames,svgnames,table]{xcolor}
\usepackage{subcaption}
\usepackage{graphicx}
\usepackage[noadjust]{cite}

\usepackage{rotating}
\usepackage{csquotes}
\usepackage{amsmath}
\usepackage{bm}
\usepackage{tabularx}
\usepackage{stfloats}
\usepackage{threeparttable}
\usepackage{url}
\usepackage{soul}
\usepackage{threeparttable}
\soulregister\cite7
\soulregister\ref7
\soulregister\pageref7
\usepackage{amssymb}
\usepackage{soul}
\usepackage{footnote}
\usepackage{colortbl}
\usepackage{soul}
\usepackage{multirow}
\usepackage{pifont}
\usepackage{algorithmic}
\usepackage{booktabs,xcolor,colortbl}
\usepackage{color}
\usepackage{alltt}
\usepackage{hyperref}
\usepackage{enumerate}
\usepackage{siunitx}
\usepackage{epstopdf}
\usepackage{bbding}
\usepackage{pbox}
 \usepackage{amsmath}  
 \usepackage{amssymb}   
 \usepackage{amsthm}
 \usepackage[dvipsnames]{xcolor}
 
\usepackage{algorithm}

\newcommand{\probP}{\text{I\kern-0.15em P}}
\newcommand{\probE}{\text{I\kern-0.15em E}}

\begin{document}

\title{Transparency and Privacy: The Role of Explainable
AI and Federated Learning in Financial Fraud Detection}

\author{
\IEEEauthorblockN{Tomisin Awosika, Raj Mani Shukla, and Bernardi Pranggono\\}
\IEEEauthorblockA{School of Computing and Information Science, Anglia Ruskin University, UK\\}
{Email: tma132@student.aru.ac.uk, raj.shukla@aru.ac.uk, bernardi.pranggono@aru.ac.uk}
}
\IEEEoverridecommandlockouts

\maketitle
\begin{abstract}
Fraudulent transactions and how to detect them remain a significant problem for financial institutions around the world.
The need for advanced fraud detection systems to safeguard assets and maintain customer trust is paramount for financial institutions, but some factors make the development of effective and efficient fraud detection systems a challenge. One of such factors is the fact that fraudulent transactions are rare and that many transaction datasets are imbalanced; that is, there are fewer significant samples of fraudulent transactions than legitimate ones. This data imbalance can affect the performance or reliability of the fraud detection model. Moreover, due to the data privacy laws that all financial institutions are subject to follow, sharing customer data to facilitate a higher-performing centralized model is impossible. Furthermore, the fraud detection technique should be transparent so that it does not affect the user experience. Hence, this research introduces a novel approach using Federated Learning (FL) and Explainable AI (XAI) to address these challenges. FL enables financial institutions to collaboratively train a model to detect fraudulent transactions without directly sharing customer data, thereby preserving data privacy and confidentiality. Meanwhile, the integration of XAI ensures that the predictions made by the model can be understood and interpreted by human experts, adding a layer of transparency and trust to the system. Experimental results, based on realistic transaction datasets, reveal that the FL-based fraud detection system consistently demonstrates high performance metrics. This study grounds FL’s potential as an effective and privacy-preserving tool in the fight against fraud.
\end{abstract}

\section{Introduction}
\label{sec:introduction}
In the era of digital banking, ensuring the security and integrity of financial activities has become paramount. While this transformation offers unparalleled convenience, it also exposes users to the vulnerabilities of cyber threats, a significant one being bank account fraud. Financial frauds, particularly in online banking and credit card transactions, pose serious threats to the global economy, the trustworthiness of financial institutions, and the financial well-being of individuals. According to the UK Finance 2022 report, billions are lost annually due to fraudulent activities, highlighting the need for more robust detection mechanisms \cite{ukfinance2022}.

With this, financial institutions have conducted and continue to undertake rigorous research to combat and identify fraud, irrespective of its nature. Nevertheless, fraud remains complicated due to its ever-evolving tactics and diverse behaviors. A prevalent domain that is the subject of extensive research is bank-related fraud \cite{abdallah2016fraud}, which translates into bank account fraud on which this research is based. 

Bank account fraud differs from other financial deceptions in its methods, impacts, and detection challenges. Unlike credit card fraud, where unauthorized transactions can be quickly detected due to unusual spending patterns, bank account fraud can manifest itself in subtler ways, such as unauthorized funds transfers, account takeovers, or even identity theft leading to the creation of new accounts \cite{pascual20172017}. The consequences for the victim can be long-lasting, both financially and emotionally. Understanding and mitigating these threats requires thorough research, underpinned by rich and diverse datasets. 

In the quest to develop systems that can detect bank account fraud, Machine Learning (ML) is frequently adopted because it effectively trains systems to deliver precise predictions based on data inputs. The choice of a specific machine learning algorithm is contingent upon the nature of the data and the specific type of fraud that the target is trying to identify. Data sets of bank account transactions not only hold confidential data but also display an imbalance, with fraudulent transactions less frequent than legitimate ones. Such characteristics present obstacles in devising a robust fraud detection system. Banks employ their proprietary data to train different ML models to recognize potentially fraudulent activities, reflecting a centralized ML methodology. This centralized approach is predominant in the financial sector today, credited largely to its proficiency in processing vast data volumes and discerning underlying patterns \cite{bhattacharyya2011data}. However, one challenge with the centralized model is that different banks often face diverse fraudulent patterns, which could hinder their ability to spot new fraudulent behaviors. This is where Federated Learning comes into the picture.

Federated Learning (FL) is a novel privacy-preserving approach to decentralized machine learning, \cite{rajesh2023take}, \cite{vyas2023histopathological}. It presents a potential solution to this predicament by enabling model training on local devices and only sharing aggregated updates. The core difference between the centralized method is in the realm of collaboration and data protection. While the former remains confined within the walls of one bank, the latter is a collective effort that spans multiple banks.

In the context of fraud detection, FL stands out as more than just a novel technological approach; it is identified as the indicator of a collaborative and confidential countermeasure against fraudulent schemes. This growing significance arises from its capacity to combine insights from different institutions without the need for direct data exchange. Moreover, FL focuses on sharing model updates rather than heavy data, which is both faster and more efficient and ensures that customer data is not compromised.

Additionally, the AI-based fraud detection techniques are black-box in nature and not transparent. For critical applications, like bank fraud detection, it is imperative that the AI system is accurate as well as trustworthy. To address the problem, we integrate Explainable AI (XAI) methods in the given FL-based banking fraud detection. In this regard, the proposed method not only preserves user privacy and provides a collaborative infrastructure to train AI models, but is also trustworthy. Thus, in this research, a fraud detection technique is proposed that uses the combined strengths of FL and XAI is proposed. The standout benefits of this methodology are numerous, with a central emphasis on user privacy preservation and transparency.  The following is a summary of the contributions of this study.
\begin{enumerate}
    \item Incorporate an FL approach for an advanced fraud detection system to ensure that individual data remains localized allowing only model updates to be centralized thus enhancing the privacy of banking datasets. This design inherently bolsters the privacy of banking datasets, a crucial advantage in our data-sensitive age. 
    \item  Develop a Deep Neural Network (DNN) model tailored to recognize patterns associated with fraudulent activities across federated databases, ensuring high accuracy. 
    \item Integration of the XAI technique to provide transparency in the model’s decisions, ensuring a novel approach to fraud detection systems.
    \item Integrating the proposed FL-based system into a web-based application to visualize the practicability of the proposed approach.
\end{enumerate}

The following are the remaining sections of this paper. Section \ref{sec:relatedwork} covers the relevant literature review and related work. In Section \ref{sec:method}, we present our methodology. We explain our detailed implementation in Section \ref{sec:implement}. In Section \ref{sec:eval}, we discuss our results. The research is concluded in section~\ref{sec:conc}.

\section{Related Work}
\label{sec:relatedwork}
Fraud detection, a very old challenge, has seen immense evolution over the years in response to the development of technology and the schemes and strategies employed by fraudsters \cite{bolton2002statistical}. 
While fraud has been a disturbing menace from ancient times, the advancement of new technology has amplified the avenues for fraudulent behavior. Technological advancements, such as communication platforms and digital finance tools, that are meant to benefit us can unintentionally also give an advantage to malicious individuals whose main goal is to cause harm. This has resulted in the rise of new types of fraud, such as mobile telecommunications scams and computer breaches.

Understanding the nature and underpinnings of fraudulent behavior has been a subject of significant scholarly attention \cite{van2018financial}. A foundational approach to this study has been the ‘fraud triangle’, a conceptual model that demystifies the factors that lead an individual to commit fraud. The authors emphasize three key elements: the incentive or pressure driving the fraudulent act, the presence of an opportunity, and the rationalization or justification by the perpetrator. Expanding on this classic model, \cite{trompeter2013synthesis} added depth by introducing three more elements: the actual act of fraud, the methods employed to conceal it, and the subsequent ‘conversion’, where the fraudsters benefit from their deceptive actions.


As financial institutions grapple with a staggering volume of fraudulent transactions, innovative solutions rooted in machine learning and deep learning have emerged at the forefront to identify and mitigate these risks.  At its core, machine learning is a subset of artificial intelligence that blends computer algorithms with statistical modeling \cite{raghavan2019fraud}. This synthesis allows computers to perform tasks without explicit programming. Instead, the system learns from the training data and uses the experiential knowledge stored to make predictions or take actions. Within machine learning’s orbit is deep learning, which harnesses artificial neural networks to decipher more intricate relationships in data. The depth and intricacy of these networks, like Convolutional Neural Networks (CNN) or Restricted Boltzmann Machines (RBM), enable them to capture unique relationships across large datasets.

A variety of machine learning and deep learning methods have been explored in the academic sphere for fraud detection. For instance, a study by \cite{zareapoor2015application} delved into the efficacy of k-Nearest Neighbors (KNN), Support Vector Machines (SVM), and ensemble classifiers in detecting fraud. Their research emphasized the challenges, like the highly unbalanced data where fraudulent transactions are less than legitimate ones, and the dynamic nature of fraud, which necessitates regularly updated machine learning algorithms. Meanwhile, other researchers like \cite{randhawa2018credit} explored the application of machine learning algorithms like Random Forest and ensemble models like AdaBoost. Sharma et al. in \cite{sharma2022credit} discussed the application of Auto-Encoders in the fraud detection framework. Auto-Encoders are specialized neural networks designed for data encoding. They operate by compressing input data into a compact representation and subsequently reconstructing it. Any significant reconstruction error, especially in a model trained on legitimate transactions, can flag potential anomalies. Parallelly, the Restricted  Boltzmann Machine (RBM) can learn a probability distribution over its set of inputs. The RBM’s ability to detect intricate patterns in unlabeled data makes it apt for identifying unauthorized transactions in vast, imbalanced datasets where fraudulent activities are but a minuscule fraction \cite{pumsirirat2018credit}.

Historically, many ML-based approaches to fraud detection have been centralized. In a centralized learning system, individual assets or clients transmit their data to a central hub or server, where data management and training occur \cite{kamei2023comparison}. However, this centralized approach presents challenges, especially in industries. The transfer of confidential data poses risks related to latency, data security, and privacy issues. In many cases, data owners might be unwilling or legally restricted from sharing sensitive information, complicating the process.

In response to these challenges, the focus has shifted to decentralized learning approaches. FL stands out as a promising solution \cite{mcmahan2017communication}. Originally developed for cellular phones, FL offers a mechanism in which the end user's data remains on their device, ensuring data privacy. In this approach, instead of sending raw data, only model updates or changes are sent to the central server, which aggregates these updates to refine the global model  \cite{mcmahan2017communication}. Notably, FL is adaptable and can handle both homogeneous independent and identically distributed (IID) and heterogeneous nonindependent identically distributed (non-IID) data. This flexibility is especially beneficial when assets have different failure modes or operate under diverse conditions. 

However, the choice between centralized and decentralized models does not only depend on data security and privacy concerns. The effectiveness of the chosen machine learning model also plays a pivotal role in fraud detection. Although recurrent neural networks (RNNs), especially long-short-term memory networks (LSTMs), have been popular for such predictions, they have inherent limitations. For instance, LSTMs, due to their sequential nature, tend to forget long sequences from earlier time-steps \cite{benchaji2021credit}. 

The choice between centralized and decentralized machine learning systems in finance and asset management is multifaceted. While the centralized approach offers simplicity and direct control, it poses significant challenges related to data transfer, privacy, and security. On the other hand, decentralized systems, especially those based on FL, provide a more private and potentially efficient approach. As industries continue to evolve and prioritize operational reliability and safety, the balance between these two systems will likely shift, with more innovations in decentralized learning approaches leading the~way.


FL presents a potential solution to this predicament by enabling model training on local devices and only sharing aggregated updates. FL’s core idea is to train machine learning models across multiple decentralized devices or servers that hold local data samples, without the need to exchange the data itself. The central server aggregates these updates to form a global model, which is then sent back to each device. This global model is refined with more rounds of localized training and aggregation, leading to a model that benefits from all available data without directly accessing it \cite{bharati2022federated}. 
A study by \cite{yang2019ffd} details the growing concern about data privacy that creates obstacles for banks to share data. Concurrently, most fraud detection systems are developed internally, keeping model details confidential to maintain data security. 

To address the problem, in this paper, we propose a federated learning-based architecture for banking fraud detection. Furthermore, we integrate the XAI techniques into the proposed FL-based method to impart the additional advantages of transparency and trust to create a more robust financial system. In addition to developing a theoretical infrastructure for the FL platform, we also deploy it using a web-based framework.

\section{Methodology} \label{sec:method}
In this section, we explain the proposed system architecture and the methodology adopted in this research.
\subsection{System architecture}
\begin{figure}[h]
	\centering
	\includegraphics[width=1\linewidth] {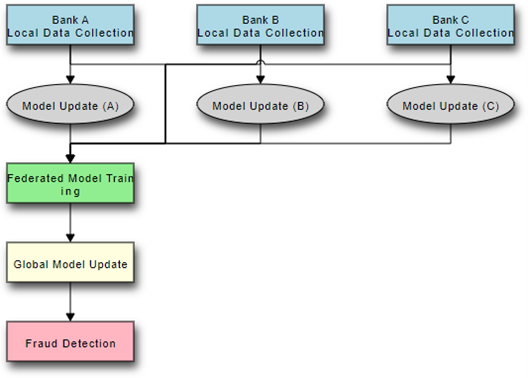}
	\caption{Proposed Federated Learning Architecture}
	\label{fig:FLArch}
\end{figure}

The proposed architecture diagram of the system shown in Fig. \ref{fig:FLArch} outlines the structure and flow of the FL approach for detecting bank fraud. Central to this architecture is a server that orchestrates the coordination of model training and aggregation across numerous banks and financial institutions. Each institution functions as a distinct node, housing its own local DNN model trained on proprietary data, ensuring that data never leaves its premises, thereby bolstering data privacy. Through periodic communication, these local models transmit their insights, not the data itself, to refine and improve a global model. This architecture not only capitalizes on the collective intelligence of all participating entities but also respects the imperative need for data security in the financial domain. 

The client and the server both have their individualistic roles and actions that they perform. The server initializes the global model, often starting with random weights or from a pre-trained model, and distributes this model to all participating clients (banks) for local training. Once clients complete their local training and send back model updates, the server is responsible for aggregating these updates to refine the global model. This typically involves averaging weights, but more sophisticated aggregation algorithms can also be employed. After aggregation, the server may validate the newly updated global model using a held-out validation set to ensure its performance meets the required standards. The server oversees the synchronization of model updates, ensuring that clients are working with the most recent version of the global model. It also manages any necessary communication between clients, although direct client-to-client communication is typically minimal in FL. 

Clients on the other hand are responsible for training the received global model on their local dataset. This involves running several training epochs to refine the model based on their specific data. After local training, clients send the model updates (e.g., weight changes) back to the central server. This does not involve sending any raw data, thus preserving data privacy. Clients ensure that raw data never leaves their local environment. All data preprocessing, cleaning, and training are done in-house, ensuring data confidentiality and compliance with privacy regulations. After the server aggregates updates and refines the global model, clients receive this updated model for subsequent rounds of training. Using the locally trained model, clients can perform real-time fraud detection on new transactions, leveraging the insights from the collective intelligence of the FL system without compromising data security. This approach not only guarantees the consistency of the globally shared model but also ensures its convergence. Instead of using generic models, a fine-tuned Deep Neural Network (DNN) model was specifically designed to recognize patterns associated with fraudulent activities across federated databases. The architecture of the model capitalizes on the immense computational strength of DNNs to discern and pinpoint fraudulent activities more effectively.



Furthermore, the proposed system model combines FL with XAI to bring forth a novel approach to fraud detection. While FL ensures efficient model training across various devices without compromising data privacy, XAI offers transparent and interpretable model decisions. This combination is pivotal, especially in sectors where understanding the rationale behind a model’s decision is as critical as the decision itself \cite{doshi2017towards}. With FL, the benefits are manifold. First and foremost, data privacy and security are significantly enhanced, as raw data remains at its origin, mitigating the risk of breaches during transfers to centralized servers — a pivotal safeguard considering the vulnerability of sensitive financial data to cyber-attacks \cite{bonawitz2019towards}. Furthermore, FL capitalizes on efficient data utilization by training models on real-time, varied data from multiple origins, leading to a more comprehensive and current fraud detection system \cite{yang2019federated}.

\begin{figure}
	\centering
	\includegraphics[width=1\linewidth]{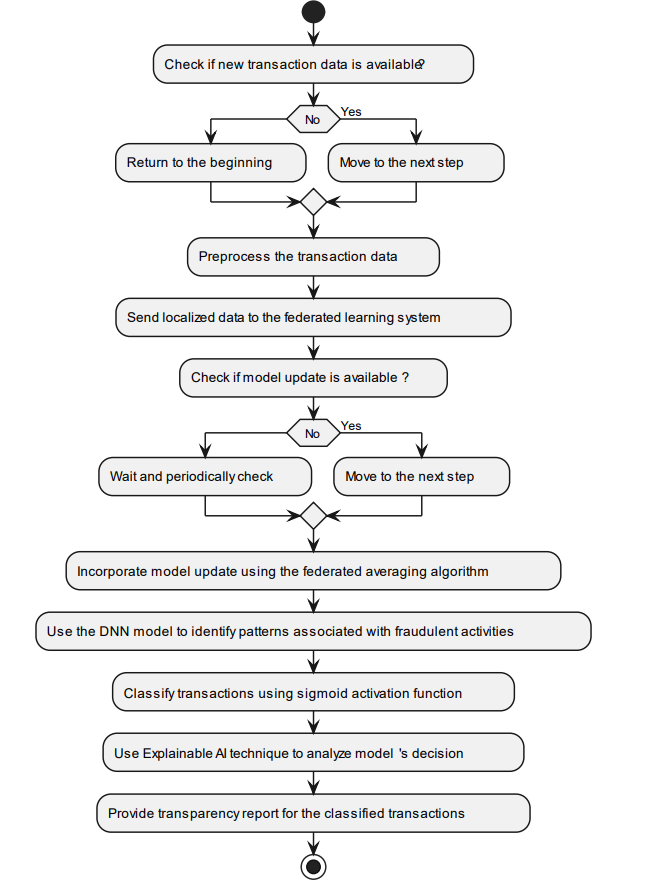}
	\caption{Workflow of the proposed System}
	\label{fig:WorkFlow}
\end{figure}

\subsection{Proposed Federated Learning-based model}
Utilizing FL for fraud detection not only leverages the power of collective data without compromising individual data privacy but also promotes more collaborative efforts among institutions to combat fraud in an ever-evolving landscape. A typical ML model update in a centralized setting, using Stochastic Gradient Descent (SGD), can be represented as: 

\[W_t+_1 = W_t - \eta\nabla L(W_t)\]

where \(W\) represents the model parameters, \(\eta\) is the learning rate, and $\nabla L(W_t)$ is the gradient of the loss function \(L\) with respect to the model parameters at iteration \(t\). 

In FL, this update is not done centrally. Instead, each client (device or server) computes its update, and these are aggregated in some way to update the global model. The core idea behind the Federated averaging algorithm proposed by McMahan et al. \cite{mcmahan2017communication} is to modify the standard SGD by computing several updates on each client and then averaging these updates on the server. 

For a given global model $w$, each client $k$ computes its update from its local data: 

\[W^k_t+_1 = W_t - \eta\nabla L_k(W_t)\]

Where \(L_k\) is the local loss on client \(k\). After each client has computed its local update, the server aggregates these updates to form the global model update. This aggregation in Federated averaging is typically a weighted sum of the local updates: 

\[W_t+_1 = \sum_k \frac{\eta_k}{\eta} W^k_t+_1\]

Where \(n_k\) is the number of data points on client \(k\), and \(n\) is the total number of data points across all clients.

This process repeats for several rounds until convergence. A key advantage is that only the model updates (and not the raw data) are communicated, which helps in maintaining data privacy. In essence, Federated averaging offers a compromise between local and centralized learning, allowing models to benefit from diverse local data sources without compromising user privacy.

\subsection{Explainable AI integration}
In the sphere of finance, interpretability is a necessity. The decisions and predictions made by models have profound real-world implications, so understanding these decisions is paramount. XAI has emerged to bridge this gap between the opaque nature of certain models and the requirement for transparency. These XAI techniques, when integrated into the FL model, empower stakeholders with insights, enhancing confidence in the model’s decisions. Moreover, understanding feature importance aids in model debugging and refinement, addressing potential pitfalls or biases.



One popular XAI technique is SHAP (SHapley Additive exPlanations) \cite{lundberg2017unified}. Originating from game theory, SHAP values provide a unified measure of feature importance by attributing the difference between the model’s prediction and the average prediction to each feature. For a given instance and feature, the SHAP value is the average contribution of that feature to all possible combinations of features.

The SHAP value for feature \( j \) is calculated as:
\begin{equation*}
    \phi_j(f) = \sum_{S \subseteq N \backslash\{j\}} \frac{|\mathcal{S}|! (|N| - |\mathcal{S}|-1)!}{|N|!} [f(\mathcal{S} \cup \{j\}) - f(\mathcal{S})]
\end{equation*}
where:
\begin{itemize}
    \item \( \mathcal{N} \) is the set of all features.
    \item \( \mathcal{S} \) is a subset of \( \mathcal{N} \) without feature \( j \).
    \item \( f(\mathcal{S}) \) is the prediction of the model for the input features in set \( \mathcal{S} \).
\end{itemize}


\section{Implementation details} \label{sec:implement}
This section presents the details of the implementation of the proposed technique as shown in Fig. \ref{fig:WorkFlow}. The steps involve preliminary checks, data processing, FL development, and XAI integration techniques.

\subsection{Dataset}
The data set referenced in this paper is sourced from \cite{jesusTurningTablesBiased2022}, encompassing realistic data based on a present-day real-world dataset for fraud detection. The dataset contains 29,042 entries, spread across 32 distinctive features. This rich dataset incorporates various data types: 17 columns of integer type, 10 columns representing floating-point numbers, and 5 columns containing categorical or string data.

\begin{figure}
	\centering
	\includegraphics[width=1\linewidth] {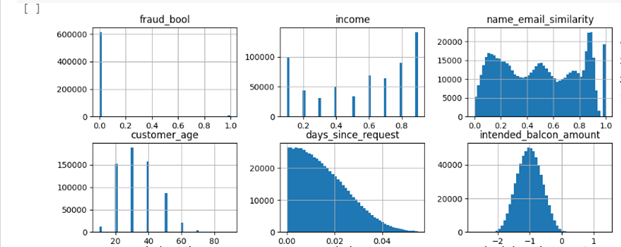}
	\caption{Characteristics of the dataset}
	\label{fig:dataset_distribution_1}
\end{figure}

\begin{figure}
	\centering
	\includegraphics[width=1\linewidth] {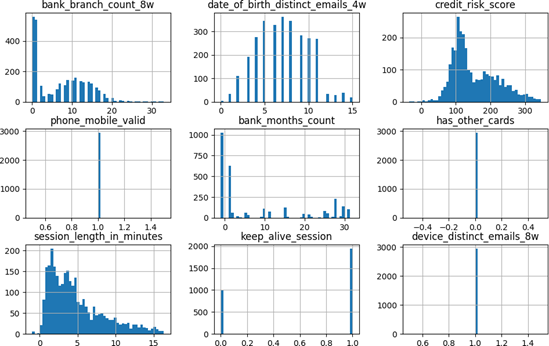}
	\caption{Characteristics of the dataset}
	\label{fig:data_visualization_2}
\end{figure}

\begin{figure}
	\centering
    \includegraphics[width=6cm]{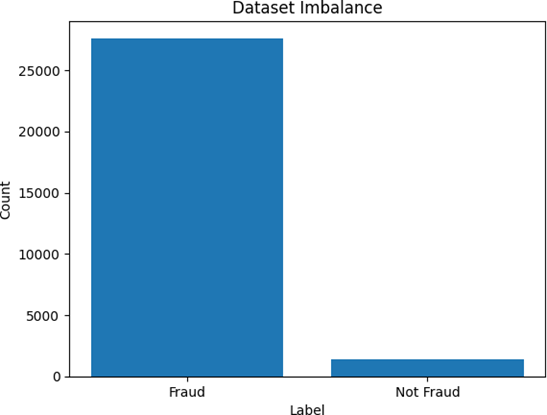}
	\caption{Imbalanced distribution of the proposed dataset}
	\label{fig:dataset_distribution}
\end{figure}

Some of the  dataset features are as follows:
\begin{enumerate}
    \item fraud\_bool: This is a binary feature and target variable for predictive models. It is an indicator of whether the record was fraudulent (1) or not (0).  
    \item income: Represents the income of the user and is a continuous variable of type float. 
    \item name\_email\_similarity: A continuous variable, capturing the similarity score between the name and email. 
    \item prev\_address\_months\_count, current\_address\_months-\_count: Indicators of the duration (in months) at the previous and current addresses. 
    \item customer\_age: Age of the customer. 
    \item days\_since\_request: A continuous variable that represents the number of days since a particular request (maybe a credit request) was made. 
    \item intended\_balcon\_amount: The amount on the balcony or a credit amount. 
    \item payment\_type, employment\_status, housing\_status, source,  device\_os: These are categorical features indicating the method of payment, the employment status of the customer, their housing situation, where the data came from, and the operating system of the device used, respectively. 
    \item credit\_risk\_score: A continuous variable possibly indicating the riskiness of providing credit to the individual or entity. 
    \item email\_is\_free: A binary variable indicating whether the email provider is free (like Gmail, Yahoo) or not. 
    \item phone\_home\_valid,  phone\_mobile\_valid: Binary or score-based indicators denoting the validity of home and mobile phone numbers. 
    \item month: This is probably indicating the month when the data was recorded or the transaction occurred. 
\end{enumerate}

The distribution of the dataset's features is shown in Fig. \ref{fig:dataset_distribution_1} and Fig. \ref{fig:data_visualization_2}. 

\subsection{Data balancing}
Since the data was highly imbalance as depicted in Fig. \ref{fig:dataset_distribution}, balancing was performed using the Synthetic Minority Over-sampling Technique (SMOTE) \cite{chawla2002smote}. The SMOTE algorithm is a popular technique to address class imbalance by generating synthetic samples in the feature space.  It is used to ensure that both classes (majority and minority) have an equal number of samples, thus addressing any imbalance present in the training data. 

\subsection{Data Pre-Processing}
In the conducted analysis, missing data in the dataset was managed using two distinct strategies. For numerical attributes, the mean value of the respective column was used to input the missing entries. In contrast, for categorical attributes, the mode, or most frequently occurring value, was employed for imputation purposes.

In the data preprocessing stage, outlier removal was undertaken for columns containing floating-point values. Utilizing the Interquartile Range (IQR) technique, any data values that were beyond 1.5 × IQR from the first (Q1) or third quartile (Q3) were identified as outliers and consequently excluded from the dataset to enhance the data’s robustness \cite{wan2014estimating}.

\begin{figure*}
	\centering
	\includegraphics[width=0.75\linewidth]{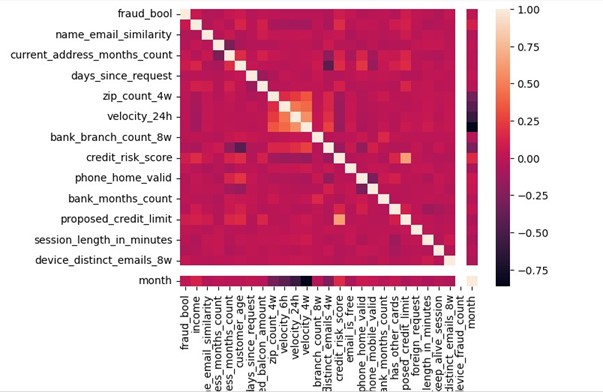}
	\caption{Fraud Dataset Correlation Matrix}
	\label{fig:DataCorrMatrix}
\end{figure*}

Subsequently, to delve into the relationships and potential dependencies among the numeric attributes, a correlation matrix was computed, as depicted in Fig. \ref{fig:DataCorrMatrix}. This matrix was then visualized as a heatmap, providing a color-coded depiction of the pairwise linear relationships among variables. The hues in this heatmap ranged from shades representing perfect negative correlation to those indicating perfect positive correlation, allowing for quick identification of strong correlations or potential multicollinearity scenarios.

\subsection{Feature Selection}
To enhance the model’s ability to learn from the data by presenting it in a more amenable format, feature engineering was performed:
\begin{enumerate}
    \item Binning: The continuous income column was binned into intervals to create the binned\_income column. Binning is a technique used to convert continuous data into discrete groups (or bins) \cite{hegland2001data}. By converting the income into ten discrete bins and labeling them with integers, the model can potentially discern patterns or trends more easily 
    across income ranges rather than individual income values. 
    \item One-hot Encoding: Columns such as employment\_status, housing\_status, payment\_type, source, and device\_os were one-hot encoded. One-hot encoding is a process by which categorical variables are converted into a format that could be provided to machine learning algorithms to do a better job in prediction. For each unique value in the original categorical column, a new binary (0 or 1) column is created. This transformation is essential for models that work better with numerical input, such as neural networks. 
\end{enumerate}

\subsection{Deep Learning model}
The deep learning (DL) model chosen for this study comprises a three-layer dense neural network. The initial layer contains 64 neurons and employs the ReLU activation function, taking its input shape from the validation dataset’s feature dimension. The subsequent layer, also activated by ReLU, has 32 neurons. The terminal layer, designed for binary classification, consists of a single neuron activated by a sigmoid function. The model’s weights are adjusted using the Adam optimization algorithm, and the binary cross-entropy function evaluates prediction losses.

\subsection{Training and Validation}
In this research, data splitting was carried out after the data had been pre-processed.  The widely accepted method of splitting the data into 80\% for training and 20\% for testing ensuring random and unbiased partitioning was carried out in this research.  To incorporate these datasets into FL, the adjusted training dataset was split into three parts: X\_train1, X\_train2, and X\_train3.


\subsection{Performance Metrics}
To effectively assess the performance of an ML model, it is crucial to employ an appropriate metric that reflects the accuracy and reliability of the model. For binary classification tasks, such as distinguishing between two outcomes, the confusion matrix is a commonly used, straightforward metric. This matrix provides four distinct prediction outcomes: (i) True Positive, where a fraudulent transaction is accurately identified as fraud; (ii) False Positive, where a legitimate transaction is mistakenly labeled as fraud; (iii) False Negative, where a fraudulent transaction is incorrectly marked as legitimate; and (iv) True Negative, where a legitimate transaction is correctly classified as such. A multi-pronged metric approach was employed to ascertain model performance, utilizing the validation dataset: 
\begin{itemize}
    \item Accuracy reflects the proportion of correct predictions made.
    \item Precision quantifies the accuracy of positive predictions.
    \item Recall highlights the fraction of positives that were rightly classified.
    \item F1-Score offers a harmonic balance between precision and recall, particularly crucial when faced with class imbalances.
\end{itemize}

\subsection{Explainable AI integration}
In terms of feature importance visualization, the SHAP method is employed, offering an interpretation of feature impact relative to a specified baseline value \cite{lundberg2017unified}.

\subsection{Simulation Setup}
Incorporating FL and using modern DL frameworks, the simulation setup provides a holistic perspective on how individual client-side models can contribute to the learning of a global model. The FL architecture aims to decentralize model training across multiple clients without sharing raw data. Post-training, using SHAP, we elucidate model decisions to make them more interpretable. The simulation was designed using a suite of software tools to facilitate FL and subsequent evaluations. Flask was employed to craft a lightweight web application framework for server-client interactions \cite{dwyer2016flask}. The TensorFlow library facilitated DL operations, particularly constructing, training, and evaluating the neural network model \cite{hope2017learning}. Cross-Origin Resource Sharing (CORS) was managed through the Flask-CORS extension, ensuring seamless AJAX cross-origin functionality. The server was configured to operate locally, accessible via port 5000.

\section{Evaluation and Results} \label{sec:eval}
In this section, the performance metrics of the federated model were well evaluated. Additionally, we also showcase the web-based framework developed in this research. Furthermore, the power of XAI was harnessed to understand the model’s decisions, identifying key features that play pivotal roles in detecting fraudulent activities. Through this exploration, the intention is to validate the effectiveness of this approach and provide insights that could reshape the landscape of fraud detection strategies.

\subsection{Web-based framework}

\begin{figure*}
    \centering
    \begin{subfigure}{0.48\linewidth}
	\includegraphics[width=\linewidth]
 {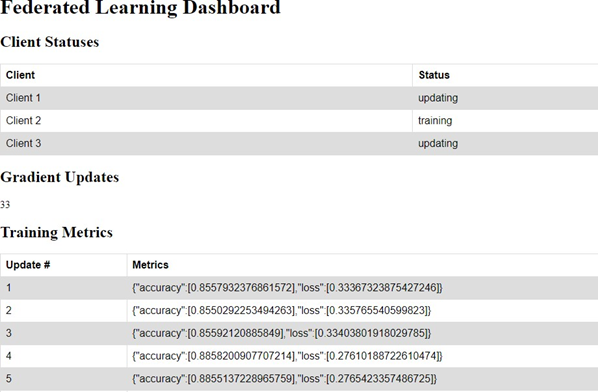}
	\caption{Federated Learning Dashboard}
	\label{fig:FLDashboard}
\end{subfigure}
\begin{subfigure}{0.48\linewidth}
	\includegraphics[width=\linewidth]{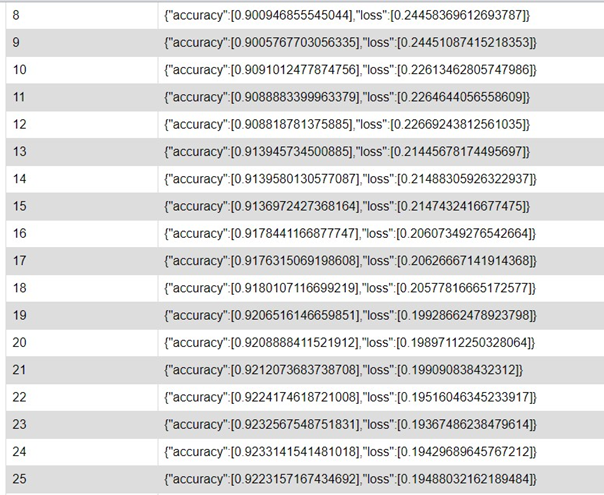}
	\caption{Federated Learning Dashboard (training process)}
	\label{fig:FLDashboard3}
\end{subfigure}
\begin{subfigure}{0.48\linewidth}
	\includegraphics[width=\linewidth]{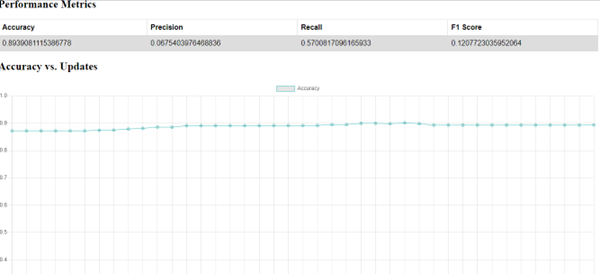}
	\caption{Federated Learning Dashboard (performance metrics)}
	\label{fig:FLDashboard2}
\end{subfigure}
\begin{subfigure}{0.48\linewidth}
	\includegraphics[width=\linewidth]{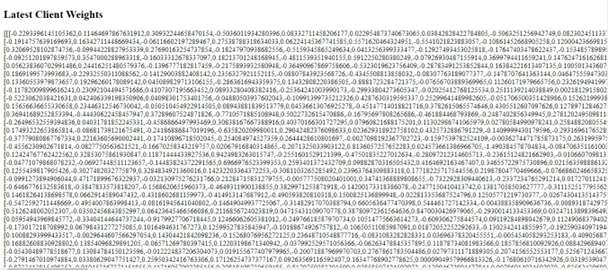}
	\caption{Federated Learning Dashboard (Weights)}
	\label{fig:FLDashboard4}
\end{subfigure}
\label{fig:dashboard}
\caption{The FL dashboard description of different pages}
\end{figure*}

Fig. \ref{fig:FLDashboard} represents the FL setup. The HTML page displays the status of each client as it trains its local dataset with the global model fetched from the central server and then updates the server with its model weights (not real data). The three status updates the clients have are: updating, training, and idle. The client status becomes idle when it is waiting for the updated global model to be sent from the server. As the server and client send updates to one another in real time, the Gradient Updates tab which shows ‘33’ is the number of times the federated learning model has been trained on the client systems. For each iteration, the accuracy, precision, recall, and F1-score are calculated and updated on the page. 

Fig. \ref{fig:FLDashboard2} depicts a real-time accuracy over updates graph. This graph also updates itself as the Gradient Update increases and more accuracy is computed. The accuracy of the FL model as shown above indicates that the model converges due to the time taken to aggregate updates from all clients. The accuracy increases exponentially and has a definite accuracy score of  93\%. Similarly, Fig. \ref{fig:FLDashboard3} and Fig. \ref{fig:FLDashboard4} show the training process and weight updates.

\subsection{Model Performance Metrics}
\begin{figure}
    \centering
\begin{subfigure}{0.45\linewidth}
	\includegraphics[width=\linewidth]{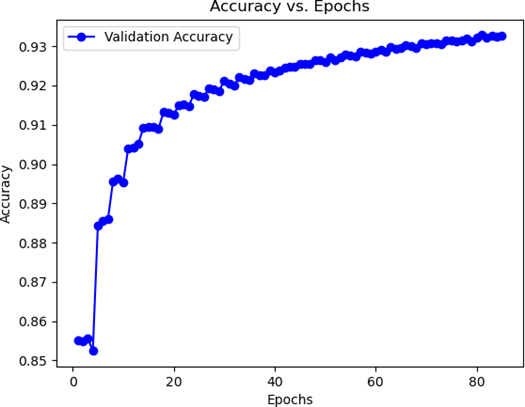}
	\caption{Accuracy of the Federated Learning Model}
	\label{fig:FLAccuracy-Epoch}
\end{subfigure}
\begin{subfigure}{0.45\linewidth}
	\includegraphics[width=\linewidth]{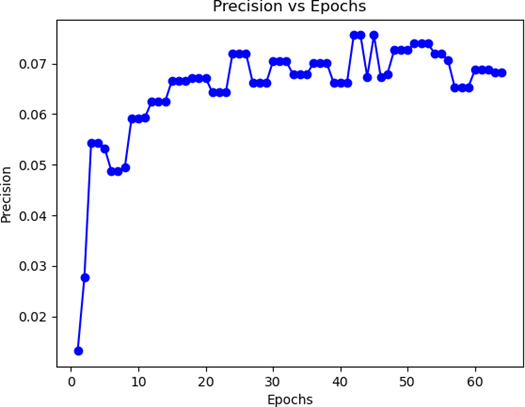}
	\caption{Precision of the Federated Learning Model}
	\label{fig:FLPrecision-Epoch}
\end{subfigure}
\begin{subfigure}{0.45\linewidth}
	\includegraphics[width=\linewidth]{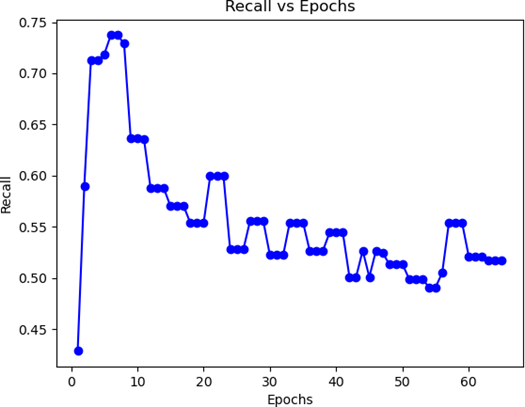}
	\caption{Recall of the Federated Learning Model}
	\label{fig:FLRecall-Epoch} 
\end{subfigure}
\begin{subfigure}{0.45\linewidth}
	\includegraphics[width=\linewidth]{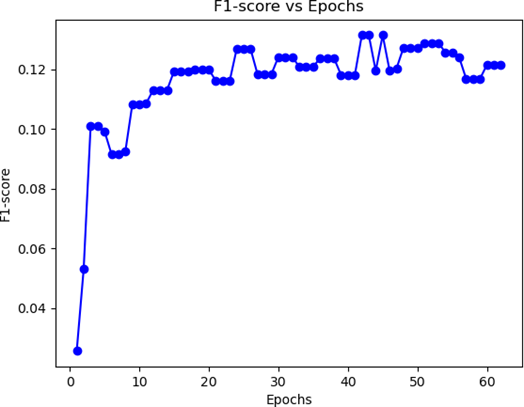}
	\caption{F1-score of the Federated Learning Model}
	\label{fig:FLF1score-Epoch}
\end{subfigure}
\caption{Performance meterices}
    \label{fig:performance meterices}
\end{figure}

The accuracy of the FL model as shown in Fig. \ref{fig:FLAccuracy-Epoch} indicates that the model converges due to the time taken to aggregate updates from all clients. The accuracy increases exponentially and has a definite accuracy score of 93\%.

Fig. \ref{fig:FLPrecision-Epoch} shows that the precision of the FL model increases linearly as the number of epochs increases until around 40 epochs. After around 40 epochs, the precision starts to fluctuate and does not increase as much. This may indicate that the model is reaching its limit of learning from the training data. Fig. \ref{fig:FLRecall-Epoch} shows that the recall of the FL model increases with the number of epochs. This indicates the model is able to learn more about the training data and improve its ability to identify positive examples over time.  Fig. \ref{fig:FLF1score-Epoch} shows that the F1-score of the FL model increases with the number of epochs. This indicates that the model is able to learn more about the training data and improve its predictions over time. However, the graph also shows that the rate of improvement decreases as the number of epochs increases. This is because the model is eventually able to learn all that it can from the training data, and any further improvement is minimal.

The research aimed to understand the benefits of using an FL-based approach for advanced fraud detection systems in terms of data privacy. FL compared to other centralized ML models showed that real data was not shared, instead, model weights were sent to a central server to perform aggregations and produce a global model. This iterative approach formed the base idea and concept of the federated learning model.


\subsection{Explainable AI Insights}
\begin{figure}
    \centering
    \begin{subfigure}{1\linewidth}
	\includegraphics[width=\linewidth]{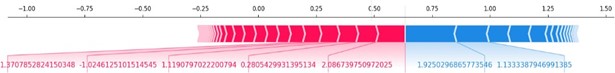}
	\caption{SHAP Plot of Client 1}
	\label{fig:SHAPclient1}
\end{subfigure}

    \begin{subfigure}{1\linewidth}
	\includegraphics[width=\linewidth]{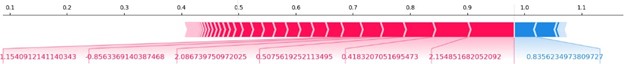}
	\caption{SHAP Plot of Client 2}
	\label{fig:SHAPclient2}
\end{subfigure}

    \begin{subfigure}{1\linewidth}
	\includegraphics[width=\linewidth]{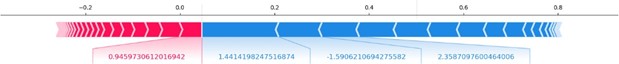}
	\caption{SHAP Plot of Client 3}
	\label{fig:SHAPclient3}
\end{subfigure}
    \caption{Explainable AI insights}
    \label{fig:xai}
\end{figure}

We have used the SHAP plots for the model explainability \cite{lundberg2017unified}.In the SHAP plots, the color coding (red and blue) represents the positive and negative impact of feature values on the prediction of the output, relative to the baseline value. The baseline value is the average of all model output values over the dataset and serves as a reference point. In the SHAP plots, the red indicates that a particular feature value represented with numbers increased the prediction value. While the blue indicates that a particular feature value decreased the prediction value. The feature 2.086 on the red side of the two client plots indicates the feature has a positive impact on the prediction of the FL model. The magnitude (that is, the distance from the baseline) of the SHAP values can indicate the strength of the influence of a feature. Higher magnitudes, whether positive or negative, signify that the feature has a strong impact on the prediction. For example, the features 2.15, 0.418, and 0.517 in the second SHAP plot (Fig. \ref{fig:SHAPclient2}) have more impact on the model prediction than 1.15 and  -0.8. 

\section{Conclusion} \label{sec:conc}
In the FL-based approach presented for banking fraud detection, a decentralized approach to model training was employed, allowing clients to train on their local data and subsequently share model updates with a central server. This approach prioritizes data privacy, as raw data remain local while still benefiting from the insights of diverse datasets. The project utilized Flask for server-client communication and handled potential scalability challenges by integrating multi-threading.  Client weights, which are multidimensional arrays representing the learned parameters of the neural network, played a crucial role in aggregating insights from individual models to update the global model. By integrating SHAP, the project not only focuses on achieving accurate model predictions but also sheds light on which features are most influential in driving these predictions. This can be especially beneficial in sensitive domains where understanding the rationale behind predictions is as important as the predictions themselves. In the context of the project, SHAP offers a pathway to build trust and ensure that the decisions of the federated model can be understood and justified.  Overall, the project highlights the efficacy and potential of FL in scenarios where centralized data collection might be impractical or undesirable due to privacy or security concerns. 
\bibliographystyle{IEEEtran}
\bibliography{reffs.bib}


\end{document}